\documentclass[11pt,a4paper]{article}
\usepackage[T1]{fontenc}
\usepackage[utf8]{inputenc}
\usepackage{textcomp}
\usepackage{lmodern}
\usepackage{graphicx}
\usepackage{amsmath,amssymb,amsthm}
\usepackage{booktabs}
\usepackage{hyperref}
\usepackage{xcolor}
\usepackage{enumitem}

\newcommand{\projectname}{MathLedger}
\newcommand{\experimentID}[1]{\texttt{#1}}

\newcommand{\valMirrorCoverage}{100.0}
\newcommand{\valBlocksAudited}{100}

\newcommand{\valHtShortHash}{\texttt{9bc8076}}

\newtheorem{definition}{Definition}
\newtheorem{proposition}{Proposition}
\newtheorem{lemma}{Lemma}
\newtheorem{remark}{Remark}

\title{\textbf{\projectname}: A Verifiable Learning Substrate with Ledger-Attested Feedback}
\author{Ismail Ahmad Abdullah\\
\texttt{ismail.abdullah.23@cnu.edu}}
\date{\today}

\begin{document}

\maketitle

\begin{abstract}
Contemporary AI systems achieve extraordinary performance yet remain opaque and non-verifiable, creating a crisis of trust for safety-critical deployment. We introduce \projectname{}, a substrate for \emph{verifiable machine cognition} that integrates formal verification, cryptographic attestation, and learning dynamics into a single epistemic loop. The system implements \emph{Reflexive Formal Learning} (RFL), a symbolic analogue of gradient descent where updates are driven by verifier outcomes rather than statistical loss.

Phase I experiments validate the measurement and governance substrate under controlled conditions. CAL-EXP-3 validates measurement infrastructure ($\Delta p$ computation, variance tracking); separate stress tests confirm fail-closed governance triggers correctly under out-of-bounds conditions. No convergence or capability claims are made. The contribution is infrastructural: a working prototype of ledger-attested learning that enables auditability at scale.

\end{abstract}

\section{Introduction: The Verifiability Gap}
\label{sec:intro}

Modern large language models are universal approximators of text, not of truth. Hallucination is structurally baked into density-estimation objectives; conventional evaluations penalize abstention and reward confident output regardless of correctness \cite{marcus2020}. In safety-critical domains---finance, law, infrastructure, policy---this creates an untenable gap between capability and trust.

The AI industry is discovering a structural constraint:
\begin{quote}
\emph{Performance without verifiability is not deployable at scale.}
\end{quote}

Mathematics offers a way out: verifiable reasoning with machine-checkable proofs. \projectname{} converts mathematics into a \emph{living protocol} for learning under formal law.

\subsection{What Problem Does This Address?}

Existing approaches to improving AI reliability fall into three categories, each with limitations:

\begin{enumerate}[leftmargin=1.5em]
\item \textbf{Reward shaping (RLHF, DPO):} Human preferences guide learning, but preferences are noisy, inconsistent, and gameable. The feedback signal is statistical, not verifiable.

\item \textbf{Verifier-guided generation:} Proof assistants check outputs post-hoc, but rejected outputs provide no structured learning signal. The verifier is a filter, not a teacher.

\item \textbf{Benchmark scaling:} Larger test sets reduce variance but do not establish correctness. Passing benchmarks does not imply understanding.
\end{enumerate}

\projectname{} takes a different approach: \emph{the verifier's outcome becomes the learning signal itself}. Every update is justified by a configured verifier outcome (pass/fail/abstain), recorded in an immutable ledger. This creates a closed epistemic loop where learning is constrained to verifier-attested outcomes.

\subsection{The Chain of Verifiable Cognition}

The system implements an end-to-end pipeline:
\[
\text{Input} \to \text{Proof-or-Abstain} \to \text{Ledger Attestation} \to \text{Dual Commitment} \to \text{Policy Update}
\]

Each component is cryptographically bound:
\begin{itemize}[leftmargin=1.5em]
\item \textbf{Proof-or-Abstain:} A configured verifier (Phase I: synthetic proxy; Phase II+: Lean kernel) validates reasoning or the system explicitly abstains. No middle ground.
\item \textbf{Ledger Attestation:} Verifier-accepted events are sealed into a monotone, append-only ledger with Merkle roots.
\item \textbf{Dual Commitment:} Both reasoning artifacts ($r_t$) and interface state ($u_t$) are committed: $H_t = \mathrm{Hash}(\texttt{EPOCH:} \| r_t \| u_t)$.
\item \textbf{Policy Update:} Reflexive Formal Learning (RFL) adjusts the policy based on verification outcomes.
\end{itemize}

This architecture enables a new primitive: \emph{learning from verifier-attested outcomes rather than statistical loss}.

\subsection{What Is Genuinely New}

\projectname{} combines three elements not commonly integrated end-to-end \cite{harrison2008,buzzard2020}:

\begin{enumerate}[leftmargin=1.5em]
\item \textbf{Ledger-attested learning signals:} Unlike reward models or human feedback, the learning signal is a cryptographically committed verification outcome.

\item \textbf{Fail-closed governance:} The system cannot silently degrade. Either verification succeeds and the outcome is admitted to the ledger, or the system abstains and logs the failure.

\item \textbf{Auditability as infrastructure:} Every update has a replayable provenance chain. Post-hoc analysis can reconstruct exactly what was learned and why.
\end{enumerate}

This paper reports Phase I experiments that validate the substrate. No capability or convergence claims are made.

\section{System Architecture}
\label{sec:architecture}

\subsection{Pipeline Overview}

The experimental methodology evaluates the \projectname{} substrate under controlled conditions. Phase I focuses on measurement validation ($\Delta p$ computation, variance tracking) and fail-closed governance verification.

\subsubsection{Test Harness}
Experiments are conducted within a dedicated FO (Feedback-Optimized) cycle harness. This harness simulates realistic operating conditions, allowing for precise control over input parameters and comprehensive capture of output metrics. The harness is designed to ensure reproducibility and provide a consistent environment for comparative analysis.

\begin{figure}[htbp]
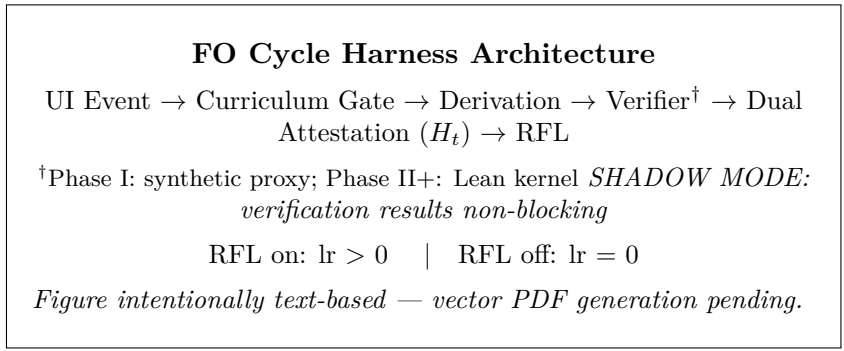

    \centering
    \fbox{\parbox{0.85\linewidth}{\centering\vspace{1em}
    \textbf{FO Cycle Harness Architecture}\\[0.5em]
    \small
    UI Event $\to$ Curriculum Gate $\to$ Derivation $\to$ Verifier$^\dagger$ $\to$ Dual Attestation ($H_t$) $\to$ RFL\\[0.5em]
    {\footnotesize $^\dagger$Phase I: synthetic proxy; Phase II+: Lean kernel}
    \textit{SHADOW MODE: verification results non-blocking}\\[0.5em]
    RFL on: lr $>$ 0 \quad|\quad RFL off: lr $=$ 0\\[0.5em]
    \textit{Figure intentionally text-based --- vector PDF generation pending.}
    \vspace{1em}}}
    \caption{Architectural overview of the FO cycle harness. The pipeline runs: UI Event $\to$ Curriculum Gate $\to$ Derivation Engine $\to$ Configured Verifier $\to$ Dual Attestation ($H_t$) $\to$ RFL policy update. Phase I uses a synthetic proxy verifier; Lean kernel integration is Phase II+. SHADOW MODE: all verification results are observational and non-blocking.}
    \label{fig:harness_architecture}
\end{figure}

\subsubsection{Configurations Under Test}
We evaluate several key configurations:
\begin{enumerate}
    \item \textbf{RFL vs. Baseline:} We compare the performance of the \projectname{} system with the RFL mechanism enabled against a baseline configuration where RFL is disabled or replaced with a static control mechanism. This comparison aims to characterize the behavior of feedback mechanisms on system stability and output.
    \item \textbf{PL slice analysis (propositional logic slices):} The system's behavior is further investigated by varying propositional-logic curriculum slices under fixed seeds and budgets. Different PL slice configurations are tested to understand their influence on computational efficiency, decision-making latency, and overall ledger integrity.
\end{enumerate}

\subsubsection{Measurement Metrics}
The following key metrics are recorded and analyzed during the experimental runs:
\begin{itemize}
    \item \textbf{Abstention Rate:} The frequency at which the system abstains from making a definitive judgment.
    \item \textbf{$H_t$ Dynamics:} We track the evolution and stability of the dual attestation hash $H_t$ over time, providing insights into the system's state processes.
    \item \textbf{Operational Metrics (runtime telemetry):} A suite of fundamental performance indicators, including transaction processing speed, error rates, and resource utilization, are measured to establish a foundational understanding of system behavior.
\end{itemize}

\subsection{The Monotone Ledger}

\begin{definition}[Monotone Ledger]
A ledger $\mathcal{L}$ is a sequence of blocks $(B_1, B_2, \ldots)$ where each $B_t$ contains verifier-accepted proof artifacts (Phase I: proxy-accepted; Phase II+: Lean-verified) with: (i) canonical statement hashes, (ii) configured verifier status, and (iii) a Merkle root $R_t$ over sorted proof IDs. The ledger is \emph{monotone} if $\bigcup_{i \le t} B_i \subseteq \bigcup_{i \le t+1} B_i$.
\end{definition}

Monotonicity ensures that accepted knowledge only grows. Statements cannot be retracted; only new proof artifacts can be added.

\subsection{Dual Attestation}

At each epoch $t$, the system commits to two roots:
\begin{itemize}[leftmargin=1.5em]
\item $r_t$: Reasoning root over canonicalized proof artifacts
\item $u_t$: UI root over interface state (DOM, logs, user confirmations)
\end{itemize}

These are bound by $H_t = \mathrm{Hash}(\texttt{EPOCH:} \| r_t \| u_t)$ with prefix-free domain separation. The tuple $(r_t, u_t, H_t)$ is the \emph{epistemic fingerprint} of the epoch---the only scalar permitted as a summary of what occurred.

\subsection{Governance-Bound Negative Knowledge}
\label{sec:negative-knowledge}

A critical requirement for fail-closed learning systems is that rejected,
failed, or inadmissible events must not disappear silently. While such events
must not influence learning or epistemic authority, they must remain
cryptographically visible for audit, replay, and governance verification.

\paragraph{Negative Knowledge as Evidence.}
\projectname{} therefore treats certain non-admitted outcomes as first-class,
audit-grade artifacts. These include:
\begin{itemize}[leftmargin=1.5em]
\item \textbf{Refuted artifacts}: reasoning attempts that were explicitly rejected
      by the configured verifier;
\item \textbf{Abstentions}: attempts that failed to satisfy verification criteria
      under bounded resources;
\item \textbf{Inadmissible updates}: learning updates blocked by active governance
      predicates or frozen commitments.
\end{itemize}

These artifacts are \emph{not} promoted to knowledge and do not enter the
monotone ledger of verified statements. Instead, they are recorded as
governance-bound evidence, preserving a verifiable record of what was
attempted and explicitly not learned. This ensures that rejected or blocked
updates constrain future interpretation and auditability without acquiring
epistemic or learning authority.

\paragraph{Frozen Governance Commitments.}
Each experimental run is bound to a versioned \emph{Governance Commitment
Registry} (GCR), whose cryptographic hash is recorded in the run manifest.
The registry enumerates non-negotiable constraints (e.g., claim ceilings,
variance bounds, update admissibility rules) that are frozen for the duration
of the run.

As a result, the system can support statements of the form:
\begin{quote}
``This update did not occur because constraint $C$ was active under
governance version $v$ for the run.''
\end{quote}

Such statements are verifiable via replay of the evidence pack and do not rely
on post-hoc interpretation or informal policy descriptions.

\paragraph{Separation from Learning Authority.}
Negative-knowledge artifacts are explicitly excluded from Reflexive Formal
Learning updates. RFL consumes only verifier outcomes summarized by
$\mathcal{V}(e) \in \{1,0,\bot\}$ as a \emph{negative signal}, never as positive
instructional content. Failed or blocked artifacts constrain admissibility but
do not teach structure.

This separation preserves three invariants:
\begin{enumerate}[leftmargin=1.5em]
\item \textbf{Soundness}: invalid reasoning cannot contaminate policy updates;
\item \textbf{Auditability}: all epistemically relevant failures remain replayable;
\item \textbf{Non-silent governance}: no event may influence future authority
      without leaving a typed, cryptographically bound trace.
\end{enumerate}

\paragraph{Relation to Dual Attestation.}
Dual attestation ($H_t = \mathrm{Hash}(\texttt{EPOCH:} \| r_t \| u_t)$) remains the
sole canonical commitment for learning and epistemic state. Governance-bound
negative knowledge operates as an \emph{orthogonal evidentiary layer}: additive,
non-authoritative, and non-interventional in Phase~I.
This evidentiary layer is replay-verified and fail-closed, but it cannot escalate claims or authorize learning updates.

\paragraph{Threat Model.}
This architecture defends against \emph{silent drift} (governance constraints
changing without trace) and \emph{policy laundering} (informal post-hoc
reinterpretation of what was permitted). It does \emph{not} protect against
malicious verifier design, compromised registry authorship, or adversarial
manipulation of the run environment itself. Phase~I assumes an honest-but-fallible
operator; Byzantine fault tolerance is out of scope.

Future phases may introduce explicit governance-state commitments, but no such
extension is required for the Phase~I claims reported here.

\section{Reflexive Formal Learning: Formal Anchor}
\label{sec:rfl}

Reflexive Formal Learning (RFL) is a symbolic analogue of gradient descent operating on verification outcomes rather than numerical errors.

\subsection{Core Definitions}

Let $\Pi$ be the space of symbolic reasoning policies and $P_\pi$ the event distribution induced by policy $\pi$.

\begin{definition}[Verification Outcome]
For reasoning event $e_t$, the verifier produces:
\[
\mathcal{V}(e_t) \in \{1, 0, \bot\}
\]
where $1$ = verification passed, $0$ = verification failed, $\bot$ = abstention.
\end{definition}

\begin{definition}[Epistemic Risk]
\label{def:epistemic-risk}
The epistemic risk of policy $\pi$ is:
\[
\mathcal{J}(\pi) = \mathbb{E}_{e \sim P_\pi}[\mathbf{1}\{\mathcal{V}(e) \neq 1\}] = \Pr_{e \sim P_\pi}[\mathcal{V}(e) \neq 1]
\]
This measures the probability mass on non-verified events (failures and abstentions).
\end{definition}

\subsection{The RFL Update Rule}

At each step $t$:
\begin{equation}
\label{eq:rfl-update}
\pi_{t+1} = \pi_t \oplus \eta_t \cdot \Phi(\mathcal{V}(e_t), \pi_t)
\end{equation}
where $\oplus$ is algebraic composition on policy space and $\Phi: \{1, 0, \bot\} \times \Pi \to \Delta\Pi$ maps verification outcomes to policy adjustments.

The intuition is:
\begin{quote}
\emph{Policies that cause fewer failures and abstentions become more likely; policies that cause them become less likely.}
\end{quote}

\begin{remark}
RFL has the mathematical structure of a stochastic approximation process (see Proposition~\ref{prop:sa-formal} in Section~\ref{sec:formal-properties}). This does not claim convergence in finite time or under Phase I conditions; convergence requires additional stability assumptions that are not claimed here. Full proofs appear in Appendix~\ref{app:formal-proofs}.
\end{remark}

\subsection{Abstention as First-Class Outcome}

Unlike reward-based systems that penalize abstention, RFL treats it as informative:
\begin{itemize}[leftmargin=1.5em]
\item Abstention prevents false positives (hallucinations committed to ledger)
\item Abstention rates provide signal about policy quality
\item High abstention with stable $\mathcal{J}(\pi)$ indicates the policy is appropriately cautious
\end{itemize}

This inverts the standard incentive structure: the system is rewarded for knowing what it does not know.

\section{Formal Properties of the Substrate}
\label{sec:formal-properties}

To strengthen the theoretical rigor of Phase I, we state formal results for three key properties: (1) the RFL update rule as a stochastic approximation process, (2) the monotonicity and tamper-evidence of the ledger, and (3) the binding property of the dual attestation hash. Each result is stated under clear assumptions; full proofs appear in Appendix~\ref{app:formal-proofs}.

\subsection{RFL Update as Stochastic Approximation}

\begin{proposition}[RFL as Stochastic Approximation]
\label{prop:sa-formal}
Consider the RFL policy update $\pi_{t+1} = \pi_t \oplus \eta_t \Phi(\mathcal{V}(e_t), \pi_t)$, where $\Phi(\mathcal{V}(e_t), \pi_t)$ is the adjustment induced by verification outcome $\mathcal{V}(e_t) \in \{1, 0, \bot\}$ at time $t$, and $\eta_t > 0$ is the learning step size. Assume $\Pi$ embeds locally into a normed vector space (or admits a coordinate chart) so that the additive form below is well-defined. Additionally assume:
\begin{enumerate}[leftmargin=1.5em]
\item \textbf{(Bounded updates)} There exists $L < \infty$ such that $\|\Phi(\mathcal{V}(e), \pi)\| \le L$ for all events and policies.
\item \textbf{(Martingale noise)} The update deviations $M_{t+1} := \Phi(\mathcal{V}(e_t), \pi_t) - h(\pi_t)$ satisfy $\mathbb{E}[M_{t+1} \mid \mathcal{F}_t] = 0$ with bounded variance, where $h(\pi) := \mathbb{E}[\Phi(\mathcal{V}(e), \pi) \mid \pi]$.
\item \textbf{(Robbins--Monro stepsizes)} $\sum_{t=0}^\infty \eta_t = \infty$ and $\sum_{t=0}^\infty \eta_t^2 < \infty$.
\end{enumerate}
Under these conditions, working in the local coordinate chart where $\oplus$ corresponds to vector addition, the RFL recursion can be written in canonical stochastic approximation form:
\[
\pi_{t+1} = \pi_t + \eta_t \big( h(\pi_t) + M_{t+1} \big)
\]
where $M_{t+1}$ is a martingale-difference noise term. By classical stochastic approximation theory \cite{Robbins1951,Kushner2003,Borkar2008}, this establishes that RFL has the mathematical structure of a learning algorithm. Convergence to an equilibrium requires additional stability assumptions (e.g., contraction of $h$) that are not claimed in Phase I.
\end{proposition}

\subsection{Monotone Ledger and Tamper-Evidence}

\begin{proposition}[Monotonicity and Tamper-Evidence]
\label{prop:ledger-formal}
Let $\mathcal{L} = (B_1, B_2, \ldots, B_T)$ be a ledger of sequential blocks, where each block $B_t$ contains verifier-accepted proof artifacts. Define the knowledge state $K_t := \bigcup_{i=1}^t B_i$. Let $L_t$ denote the ledger head hash after block $t$, computed as $L_t = \mathrm{Hash}(L_{t-1} \| R_t)$ where $R_t$ is the Merkle root of $B_t$. Assume:
\begin{enumerate}[leftmargin=1.5em]
\item Blocks are append-only (no modification after appending).
\item The hash function is collision-resistant.
\end{enumerate}
Then:
\begin{enumerate}[leftmargin=1.5em]
\item \textbf{(Monotonicity)} $K_t \subseteq K_{t+1}$ for all $t$. Accepted knowledge only grows.
\item \textbf{(Tamper-Evidence)} For any altered ledger $\tilde{\mathcal{L}} \neq \mathcal{L}$, the head hash $\tilde{L}_T \neq L_T$ except with negligible probability.
\end{enumerate}
\end{proposition}

\subsection{Dual Attestation Binding}

\begin{lemma}[Binding Property of Dual Attestation]
\label{lem:binding}
At each epoch $t$, the system commits to reasoning root $r_t$ (32-byte digest) and UI root $u_t$ (32-byte digest), then publishes $H_t = \mathrm{Hash}(\texttt{EPOCH:} \| r_t \| u_t)$. Under the assumption that the hash function is collision-resistant and the encoding uses fixed-width (32-byte) digests with prefix-free domain separation, the hash $H_t$ binds the pair $(r_t, u_t)$: it is computationally infeasible for any $(r'_t, u'_t) \neq (r_t, u_t)$ to produce the same $H_t$.
\end{lemma}

\begin{remark}
The fixed-width encoding (32-byte digests) eliminates concatenation ambiguity. The prefix \texttt{EPOCH:} provides domain separation from other hash uses in the system. Together with Proposition~\ref{prop:ledger-formal}, this ensures every aspect of the system's state is tamper-evident and auditably linked to verifier-accepted proof artifacts.
\end{remark}

\section{Phase I Experimental Results}
\label{sec:results}

This section presents Phase I experimental findings. The primary goal is validating measurement infrastructure and fail-closed governance, not demonstrating capability or convergence.

\subsection{CAL-EXP-3: Measurement Validation}

CAL-EXP-3 validates that $\Delta p$ (success rate proxy) is computable per cycle and that variance between experimental arms is measurable. The experiment compares:
\begin{itemize}[leftmargin=1.5em]
\item \textbf{Baseline (lr=0.0):} RFL disabled; policy static
\item \textbf{Treatment (lr=0.1):} RFL enabled; policy updated based on verifier outcomes
\end{itemize}

Both conditions exhibited oscillatory $\Delta p$ dynamics around the decision threshold. No convergence or uplift is claimed; the purpose is infrastructure validation. Full time-series plots are available in the evidence pack (ancillary material).

\subsection{Fail-Closed Governance}

Separate stress tests confirm that governance predicates trigger correctly under out-of-bounds conditions:
\begin{itemize}[leftmargin=1.5em]
\item \textbf{F5.2 (variance ratio):} Fires when inter-arm variance exceeds threshold
\item \textbf{F5.3 (windowed drift):} Fires when $\Delta p$ drift exceeds tolerance
\end{itemize}
When triggered, these predicates cap the claim level at L0 (no capability claim). This is the expected behavior for Phase I stress tests.

\subsection{Dual-Root Attestation}

The Mirror Auditor confirmed the integrity of the dual-root attestation mechanism. For the \experimentID{\valHtShortHash} snapshot, coverage was \valMirrorCoverage\%, with \valBlocksAudited{} blocks fully audited and verified.

\subsection{Interpreting Phase I Outcomes}

The Phase I results establish three facts:

\begin{enumerate}[leftmargin=1.5em]
\item \textbf{The measurement substrate works.} $\Delta p$ (success rate proxy) is computable per cycle. Variance between arms is measurable.

\item \textbf{Fail-closed governance triggers correctly.} In stress tests, F5.2 (variance ratio out of bounds) and F5.3 (windowed drift excessive) fired as expected, capping claims at L0.

\item \textbf{Non-convergence is informative, not a failure.} Phase I was designed to validate infrastructure, not demonstrate capability. The fact that fail-close triggers fired correctly \emph{is} the success condition.
\end{enumerate}

\section{Discussion: Why This Matters}
\label{sec:discussion}

Phase I experiments characterized the behavior of the \projectname{} substrate under controlled conditions. The focus was validating measurement infrastructure and fail-closed governance, not demonstrating capability.

\subsection{Comparison to Adjacent Work}

\projectname{} occupies a distinct position in the landscape of verifiable AI:

\begin{table}[htbp]
\centering
\begin{tabular}{l|c|c|c}
\textbf{Approach} & \textbf{Learning Signal} & \textbf{Auditability} & \textbf{Fail-Closed} \\
\hline
RLHF & Human preference & Low & No \\
Verifier-guided & Post-hoc filter & Medium & No \\
Proof-carrying code & None (static) & High & Yes \\
\textbf{\projectname{} (RFL)} & \textbf{Verifier outcome$^\dagger$} & \textbf{High} & \textbf{Yes} \\
\end{tabular}
\caption{Comparison of approaches to reliable AI. \projectname{} uniquely combines verified learning signals with fail-closed governance. $^\dagger$Phase I uses a synthetic proxy verifier; formal proof verification (Lean) is Phase II+.}
\label{tab:comparison}
\end{table}

\subsection{Layer-3 Infrastructure}

\projectname{} is not a proof generator or a user-facing application. It is \emph{Layer-3 infrastructure}: the flight data recorder for AI reasoning.

\begin{itemize}[leftmargin=1.5em]
\item \textbf{Layer 1 (Human):} Users pose queries, interpret results, make decisions
\item \textbf{Layer 2 (Engine):} AI models generate formal artifacts
\item \textbf{Layer 3 (Ledger):} \projectname{} provides immutable provenance and attestation
\end{itemize}

The system does not compete with proof generators; it makes their outputs trustworthy at scale.

\section{Explicit Non-Claims and Scope Boundaries}
\label{sec:non-claims}

To maintain epistemic discipline, we explicitly state what Phase I does \emph{not} establish:

\subsection{What Phase I Does NOT Establish}

\begin{itemize}[leftmargin=1.5em]
\item \textbf{Capability claims:} No claim that the system ``understands'' or ``reasons'' in any general sense.
\item \textbf{Convergence:} No claim that RFL converges under Phase I conditions. All runs failed the variance gate.
\item \textbf{Threshold validity:} Thresholds are frozen parameters, not validated optima.
\item \textbf{Generalization:} No out-of-distribution testing was performed.
\item \textbf{Real-world applicability:} Only synthetic harness data was used.
\end{itemize}

\subsection{SHADOW Mode Semantics}

All Phase I experiments operate in SHADOW mode: verification results are \emph{observational and non-blocking}. The system records what happened but does not gate production decisions. In this context, ``fail-closed'' means claim-capping and evidence-rejection, not production blocking---governance predicates cap the claim level at L0 when triggered, but do not halt execution.

\subsection{Phase Quarantine}

Phase I and Phase II are strictly separated:
\begin{itemize}[leftmargin=1.5em]
\item \textbf{Phase I:} Assumes ideal verifier, hermetic environment, synthetic data
\item \textbf{Phase II:} Tests governance stability under auxiliary perturbation (frozen but not executed)
\end{itemize}

No Phase II claims are made in this work. Phase II specification is frozen pending execution authorization.

\section{Future Work}
\label{sec:future_work}

Future work will focus on integrating RFL to observe if reflexive feedback can dampen oscillatory states in the decision boundary and achieve measurable reductions in abstention rates and improvements in convergence latency.

\subsection{Phase II Calibration}

Phase II of the calibration program addresses governance stability: specifically, whether the governance verdict (failure codes, claim level) is invariant under perturbation of auxiliary parameters not part of the frozen predicate set. The Phase II specification is frozen, but execution has not yet occurred. No claims regarding governance invariance or sensitivity are made in this work. Phase II results, when available, will be reported separately and will not retroactively modify the Phase I conclusions presented here.

\section{Conclusion}
\label{sec:conclusion}

\projectname{} demonstrates that ledger-attested learning is technically feasible. Phase I successfully established:

\begin{enumerate}[leftmargin=1.5em]
\item A working pipeline from proof generation through dual attestation to policy feedback
\item Measurement infrastructure for $\Delta p$ and variance metrics
\item Fail-closed governance that correctly triggers under out-of-bounds conditions
\item Explicit non-claims and scope boundaries that enable honest assessment
\end{enumerate}

The contribution is infrastructural, not empirical. We have built the substrate; demonstrating capability on that substrate is future work.

The system stands as proof-of-concept for a new paradigm: \emph{learning from verifier-attested outcomes}. Whether this paradigm scales to complex reasoning remains an open question. What Phase I establishes is that the question can now be asked with rigor.

\appendix

\section{Evidence Pack}
\label{app:evidence}

The evidence pack provides cryptographic verification of experimental runs. Key artifacts:

\begin{table}[h]
    \centering
    \begin{tabular}{l l}
        \toprule
        \textbf{Artifact} & \textbf{Contents} \\
        \midrule
        Evidence Manifest & File list with SHA-256 hashes (JSON) \\
        Run Metadata & Experiment configuration and timing (JSON) \\
        Governance Verdict & Claim level and predicate outcomes (JSON) \\
        \bottomrule
    \end{tabular}
    \caption{Key experimental artifacts. Exact paths and SHA-256 hashes provided in ancillary material.}
    \label{tab:manifest}
\end{table}

The complete evidence pack (run manifests, cryptographic hashes, raw $\Delta p$ time series) will be published as ancillary material with this submission.

\paragraph{Governance binding in the evidence manifest.}
In addition to file hashes, the evidence manifest records:
\begin{enumerate}[leftmargin=1.5em]
\item a SHA-256 hash of the active Governance Commitment Registry (GCR), computed over RFC~8785-style canonical JSON (keys sorted lexicographically, no whitespace, ASCII-safe encoding);
\item a per-artifact classification tag (\texttt{artifact\_kind}) with values: \texttt{VERIFIED}, \texttt{REFUTED}, \texttt{ABSTAINED}, or \texttt{INADMISSIBLE\_UPDATE}.
\end{enumerate}
The replay verifier checks these fields fail-closed: (i)~missing or invalid \texttt{artifact\_kind} enum values, (ii)~missing \texttt{commitment\_registry\_sha256} field, and (iii)~mismatched registry file hash all cause verification failure with exit code~1. This makes governance constraints and rejected updates cryptographically visible without elevating them to knowledge claims.

\paragraph{Scope disclaimer.}
The evidence pack verifies artifact integrity, determinism, and governance binding only. It does \emph{not} validate correctness, safety, alignment, or legal compliance. The governance commitments in the registry are illustrative placeholders in v0.9.x; the mechanism (hash binding) is what is being validated, not the normative content.

\paragraph{Version pinning.}
The external audit surface for Phase~I corresponds to Git tag \texttt{v0.9.4-pilot-audit-hardened}. The canonical verification command is:
\begin{verbatim}
uv run python scripts/run_dropin_demo.py --seed 42 --output demo_output/
cd demo_output && python verify.py
\end{verbatim}
Expected test vectors (SHA-256 hashes for seed=42) are documented in \texttt{docs/pilot/AUDIT\_WALKTHROUGH.md} within the tagged release.

\section{Formal Proofs}
\label{app:formal-proofs}

This appendix provides complete proofs for the formal properties stated in Section~\ref{sec:formal-properties}. Stronger convergence and robustness results under additional assumptions are developed in a separate technical companion and are intentionally excluded here to preserve Phase I scope.

\subsection{Proof of Proposition~\ref{prop:sa-formal} (RFL as Stochastic Approximation)}

\begin{proof}
The update $\pi_{t+1} = \pi_t \oplus \eta_t \Phi(\mathcal{V}(e_t), \pi_t)$ can be interpreted as an additive update in a suitable parameterization. Define $h(\pi) := \mathbb{E}[\Phi(\mathcal{V}(e), \pi) \mid \pi]$, the expected update given the current policy. Define the noise term:
\[
M_{t+1} := \Phi(\mathcal{V}(e_t), \pi_t) - h(\pi_t)
\]
By construction, $\mathbb{E}[M_{t+1} \mid \mathcal{F}_t] = h(\pi_t) - h(\pi_t) = 0$, so $M_{t+1}$ is a martingale difference adapted to $\mathcal{F}_t$. The update becomes:
\[
\pi_{t+1} = \pi_t + \eta_t \big( h(\pi_t) + M_{t+1} \big)
\]
This is the canonical Robbins--Monro stochastic approximation form. Under assumptions (bounded updates, martingale noise with bounded variance, Robbins--Monro stepsizes), standard SA theory applies. The function $h$ plays the role of the mean-field drift.

We emphasize: this establishes that RFL has SA \emph{structure}. Convergence to an equilibrium of $\dot{\pi} = h(\pi)$ requires that such an equilibrium exists and is attractive (e.g., $h$ is a contraction). These additional stability conditions are not claimed in Phase I.
\end{proof}

\subsection{Proof of Proposition~\ref{prop:ledger-formal} (Monotonicity and Tamper-Evidence)}

\begin{proof}
\textbf{(1) Monotonicity:} By definition, $K_t = \bigcup_{i=1}^t B_i$. When block $B_{t+1}$ is appended:
\[
K_{t+1} = \bigcup_{i=1}^{t+1} B_i = K_t \cup B_{t+1} \supseteq K_t
\]
Since blocks are append-only, no element of $K_t$ is removed. Thus $K_t \subseteq K_{t+1}$.

\textbf{(2) Tamper-Evidence:} Suppose an adversary produces $\tilde{\mathcal{L}} = (\tilde{B}_1, \ldots, \tilde{B}_T) \neq \mathcal{L}$ with the same head hash $\tilde{L}_T = L_T$. Let $j$ be the smallest index where $\tilde{B}_j \neq B_j$.

\textbf{Case A:} If $\tilde{B}_j$ differs from $B_j$ as a set, then Merkle root $\tilde{R}_j \neq R_j$ (deterministic construction). Given $L_j = \mathrm{Hash}(L_{j-1} \| R_j)$ and $\tilde{L}_j = \mathrm{Hash}(L_{j-1} \| \tilde{R}_j)$ (assuming prior blocks match), we have $\tilde{L}_j \neq L_j$ unless a hash collision occurs. By collision resistance, this happens with negligible probability.

\textbf{Case B:} If the sequence lengths differ (block omitted or inserted), the hash chain incorporates a different number of blocks, yielding $\tilde{L}_T \neq L_T$ by similar reasoning.

In both cases, $\tilde{L}_T = L_T$ implies a hash collision, which is computationally infeasible.
\end{proof}

\subsection{Proof of Lemma~\ref{lem:binding} (Dual Attestation Binding)}

\begin{proof}
The hash input is $m = \texttt{EPOCH:} \| r_t \| u_t$, where $r_t$ and $u_t$ are fixed-width 32-byte digests. This encoding is unambiguous: the prefix \texttt{EPOCH:} is a fixed string, and the 32-byte widths mean there is a one-to-one correspondence between pairs $(r_t, u_t)$ and input strings $m$.

Suppose $(r'_t, u'_t) \neq (r_t, u_t)$ yields the same hash:
\[
\mathrm{Hash}(\texttt{EPOCH:} \| r'_t \| u'_t) = \mathrm{Hash}(\texttt{EPOCH:} \| r_t \| u_t)
\]
Let $m' = \texttt{EPOCH:} \| r'_t \| u'_t$ and $m = \texttt{EPOCH:} \| r_t \| u_t$. Since the pairs differ and encoding is bijective, $m' \neq m$. Thus we have a hash collision, which is infeasible under collision resistance.

Therefore, $H_t$ uniquely commits to $(r_t, u_t)$. Once published, the agent cannot claim a different pair without finding a collision.
\end{proof}

\bibliographystyle{plain}
\bibliography{references}

\end{document}